\documentclass{article} 
\usepackage{iclr2027_conference,times}

\usepackage{amsmath,amsfonts,bm}

\def\eqref#1{equation~\ref{#1}}

\def\1{\bm{1}}

\DeclareMathAlphabet{\mathsfit}{\encodingdefault}{\sfdefault}{m}{sl}
\SetMathAlphabet{\mathsfit}{bold}{\encodingdefault}{\sfdefault}{bx}{n}

\usepackage{hyperref}
\usepackage{url}
\usepackage{booktabs,graphicx}
\usepackage{xspace}
\usepackage{caption}
\usepackage[dvipsnames]{xcolor}
\usepackage{booktabs}
\usepackage{caption}
\usepackage{wrapfig}
\usepackage{tcolorbox}
\usepackage{booktabs}
\usepackage{graphicx}
\usepackage[table]{xcolor}
\usepackage{subfigure}
\usepackage{subcaption}
\usepackage{wrapfig}

\usepackage{booktabs}
\usepackage[table]{xcolor}
\definecolor{headerblue}{RGB}{35,75,115}
\definecolor{madsblue}{RGB}{232,242,250}
\usepackage{listings}
\usepackage{float}
\usepackage{caption}

\lstdefinestyle{promptstyle}{
  basicstyle=\ttfamily\scriptsize,
  columns=fullflexible,
  keepspaces=true,
  breaklines=true,
  breakatwhitespace=false,
  showstringspaces=false,
  upquote=true
}
\usepackage{xcolor}
\usepackage{booktabs}
\usepackage[table]{xcolor}
\definecolor{softgreen}{HTML}{E8F5EC}
\title{ToolRACER: A Robust Agentic Conversation Emulation Resource for Agent Training and Evaluation}

\author{%
\makebox[\dimexpr\textwidth-2\tabcolsep\relax][c]{%
\begin{tabular}{@{}c@{}}
\noalign{\vskip 18pt}
Arkajyoti Chakraborty\textsuperscript{$\star$},
Aryan Tayal\textsuperscript{$\star$},
Ishika Agarwal\textsuperscript{$\diamond$}\\[2pt]
Tanner Sorensen\textsuperscript{$\star$},
Justin Chiu\textsuperscript{$\star$},
Alessandro Di Bari\textsuperscript{$\star$}\\[2pt]
Neha Gupta\textsuperscript{$\star$},
Andreas Stolcke\textsuperscript{$\star$}\\[5pt]
{\small
\textsuperscript{$\star$}Uniphore
\qquad
\textsuperscript{$\diamond$}University of Illinois Urbana-Champaign
}\\[3pt]
{\small\texttt{arkajyoti.chakraborty@uniphore.com}}
\end{tabular}%
}%
}

\iclrfinalcopy 
\begin{document}
\newcommand{\sysnbench}{ToolRACERBench\xspace}
\newcommand{\sysn}{ToolRACER\xspace}
\newcommand{\happy}{\textcolor{ForestGreen}{\textbf{Happy}}\xspace}
\newcommand{\unhappy}{\textcolor{Plum}{\textbf{Unhappy}}\xspace}
\newcommand{\impossible}{\textcolor{BrickRed}{\textbf{Impossible}}\xspace}

\maketitle

\begin{abstract}
Task-oriented conversational agents remain fragile under real world conversation scenarios as they rarely follow a predictable script, especially when users exhibit non-cooperative behavior. Existing function-calling benchmarks often emphasize successful, cooperative interactions and underrepresent adversarial conversation trajectories, thereby limiting the training resources available for developing robust agents. We present \textbf{\sysn}, a synthetic data generation pipeline that coordinates user, assistant and tool emulation models to generate and validated multi-turn interactions between a user and an agent. Using \sysn, we construct \textbf{\sysnbench} a robust multi-turn conversation benchmark spanning six domains, ranging over 55 varied personas, generating a validated corpus of 5.6K conversation trajectories, with approximately 66\% of conversations containing failure-prone conversation scenarios. We inject adversarial behaviors, producing validated conversational interaction trajectories that capture realistic, robust scenarios. We evaluate models trained on \sysnbench against internal benchmarks, as well as on function calling benchmarks such as $\tau^2$-bench, BFCLv3 and ACEBench to evaluate agentic accuracy and robustness. Models trained on \sysnbench improve end to end agentic accuracy across $\tau^2$-bench and ACEBench, demonstrating significant gains when mixed with in-domain dataset in small language models for agent capability tasks.     
\end{abstract}


\section{Introduction}

Task-oriented dialog systems are a key component of modern enterprise conversational AI, powering applications such as customer support, transaction handling, and automated service workflows \citep{bfcl, wang2026mcpbench}. Despite significant advances in natural language understanding and generation driven by large language models (LLMs), the dialog management layer in many deployed systems remains largely unchanged \citep{rabinovich-anaby-tavor-2025-robustness}. Most enterprise-grade systems that rely on finite-state dialogue managers or pre-scripted call flows to enforce fixed interaction paths can offer predictability and control, but struggle to handle deviations from expected user behavior, evolving goals, or incomplete information, often leading to brittle interactions and increased user friction. In contrast, human agents exhibit a high degree of flexibility in task-oriented conversations. They maintain rich conversational context, infer missing information when possible, and dynamically decide whether to ask clarifying questions or consult external resources. Bridging this gap between rigid automated systems and adaptive human behavior remains a core challenge in conversational AI \citep{barres2025tau2}.

To enable such robust agentic capabilities, we need high quality and diverse multi-turn data that has multiple exchanges between the user and the agent across various task-oriented trajectories, each requiring a sequence of interdependent actions and solving the user-initiated task through function calls. Building reliable AI agents that are responsive to shifts in conversational goals and tool failures require robust conversation trajectory planning. Recent training benchmarks \citep{prabhakar2025apigenmtagenticpipelinemultiturn, liu2024apigenautomatedpipelinegenerating, liu2025toolacewinningpointsllm} use multi-agent simulation to generate diverse corpora of high-quality synthetic conversations. Still, these works lack various aspects of realistic user interactions. 

In our framing, we break down the taxonomy of realistic user interactions into three types: \happy, \unhappy, \impossible. \happy paths are paths with non-adversarial users and achievable tasks. \unhappy paths are those with adversarial users, but achievable tasks. \impossible paths are those with non-achievable tasks.

\begin{figure}[t]
\centering

\begin{minipage}[t]{0.48\textwidth}
  \vspace{-5pt}
  \centering
  \includegraphics[width=\linewidth]{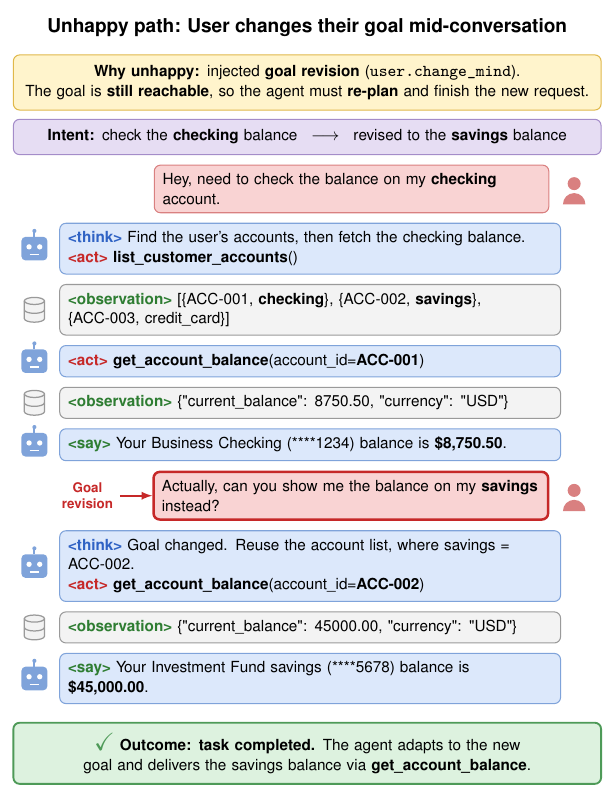}
  \par\smallskip
  {\small (a) Unhappy path.\par}
\end{minipage}%
\hfill
\begin{minipage}[t]{0.48\textwidth}
  \vspace{-5pt}
  \centering
  \includegraphics[width=\linewidth]{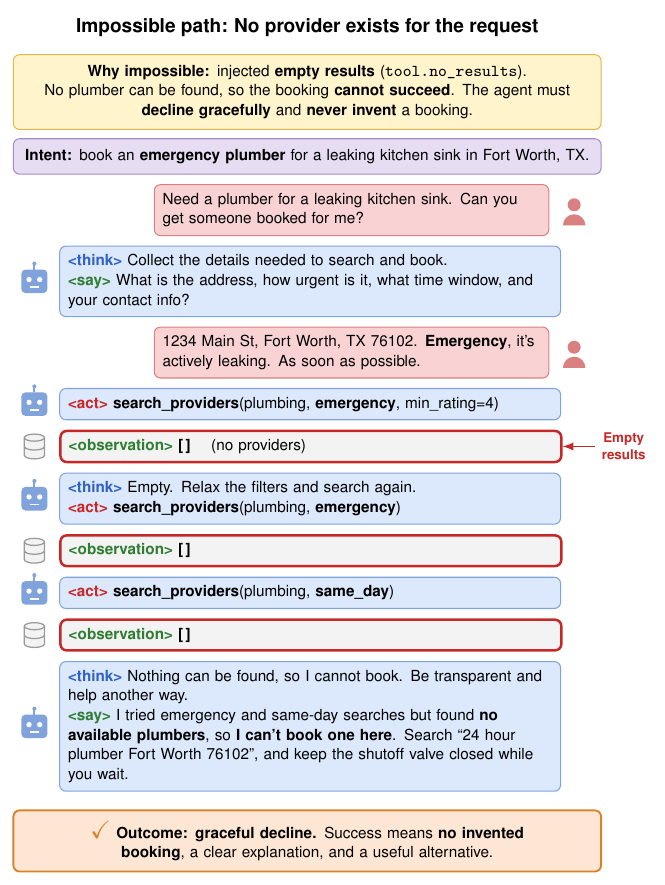}
  \par\smallskip
  {\small (b) Impossible path.\par}
\end{minipage}

\caption{Examples of \sysnbench conversation trajectories:
(a) an \unhappy path and (b) an \impossible path.}
\label{fig:path_examples}
\vspace{-15pt}
\end{figure}

To address these types of interactions, we introduce \sysn: a Robust Agentic Conversation Emulation Resource for producing high-quality multi-turn agent traces. \sysn generates diverse, validated trajectories through a pipeline consisting of three stages: 1) \textbf{Multi-agent simulation}, where three agents are initialized as user, assistant and tool agents performing dialogue simulation based on a defined set of skeleton scenarios covering \happy, \unhappy and \impossible paths, mimicking realistic world scenarios. 2) \textbf{Multi-stage evaluation}, consisting of a series of four evaluators initialized with judge LLMs that check the syntax and dialogue consistency and faithfulness of the generated traces. 3) A \textbf{Refinement loop} that improves the failed conversations from Stage 2. Stage 2 also provides failure reasons from its evaluation, which we use in Stage 3 to produce validated multi-turn tool-calling conversations. Resulting from these three stages is a benchmark that can be used for training and testing tool-calling conversational agents. We call it \sysnbench.

In summary, our contributions are as follows: 
\begin{itemize}
    \item We propose \sysn, a synthetic data generation pipeline that leverages tool execution feedback and generates robust multi-turn agentic conversations validated through a rubric of evaluators for realistic and diverse synthetic data generation. 

    \item We propose a taxonomy of adversarial trajectories, generating robust agentic traces by explicitly modeling both user-originated and environment-originated failures, classifying traces into \happy, \unhappy and \impossible scenarios. 

    \item We perform extensive experiments using \sysnbench comparing against the SOTA training benchmark APIGen-MT \cite{prabhakar2025apigenmtagenticpipelinemultiturn} and provide empirical evidence showcasing performance gains across cross domain benchmarks such as $\tau^2$-bench, BFCLv3 and ACEBench. Along with cross-domain evaluation, we evaluate ablations of our training data mixtures on the \sysnbench-test set, thus analyzing agentic performance under robust interactions. 
\end{itemize}

\section{Related Work}

\paragraph{Function-calling benchmarks}
Since the inception of BFCL (Berkeley Function Calling Leaderboard) \citep{bfcl} and $\tau$-bench \cite{yao2025taubench}, researchers have been trying to improve the conversational function-calling abilities of LLMs. API-Bank \citep{apibank} is one such early work that shows language models can improve their tool-calling abilities with curated data, but suffer from poor API-formatting and an inability to follow instructions. Subsequent tool-calling benchmarks became more realistic: having long-context conversations (BFCL V3), evaluating tool-calling of agents that need to talk to other agents (via MCP protocol) \citep{wang2026mcpbench}, and/or allowing both user and agent to act on the same environment \citep{barres2025tau2}. These methods, while they reflected the abilities of real-world environments, did not fully capture the unpredictable nature of real-world interactions.

\textbf{Robust Function-Calling Agents}
Agents would be useless in the real world if they are unable to adapt to unseen scenarios \citep{rabinovich-anaby-tavor-2025-robustness}. To make agents more robust, many benchmarks now include noise in their data, such as HammerBench \citep{hammerbench} (includes irrelevant functions), NoisyToolBench \citep{noisytoolbench} (includes ambiguous or incomplete instructions), and Multi-Mission Tool Bench \citep{yu2025multimissiontoolbenchassessing} (changes the goal during the interaction). Though agentic and environmental robustness are important, still lacking in these works is user-centric robustness.

\textbf{Trajectory-Driven Robustness}
Human users can be quite difficult. They can be uncooperative, aggressive, or impatient; they can be clueless and lost; they can also demand things that go against agent policy. However, many function-calling benchmarks assume an ideal user \citep{shi2026tautaubenchenvironmentendtoendrealistic, barres2025tau2}. Benchmarks that stress-test a system are required to ensure that a system is ready to handle real users \citep{purwar2026mmtaup2personaadaptivepromptingrobust}. Persona-driven robustness is a very common and easy-to-implement technique to create such a benchmark. For example, \citet{tan2025personabenchevaluatingaimodels} creates a variety of personas and evaluates whether they can be maintained in a RAG-style evaluation setup. \citet{chopra2026cooperativesimulatorsgeneratingrealistic} use personas in training for tau-bench tasks to show that models can be made more robust. Finally \citet{wang2026mcppersonabenchmarkingllmagents} uses personas to diversify their benchmark for evaluating personalization. Our work likewise makes use of diverse personas and situations to increase robustness in training and diversity in evaluating  tool-use scenarios.

\section{\sysn: Methodology for Robust Synthetic Multi-Turn data Generation}

\begin{figure}[t]
    \centering
    \includegraphics[width=\linewidth]{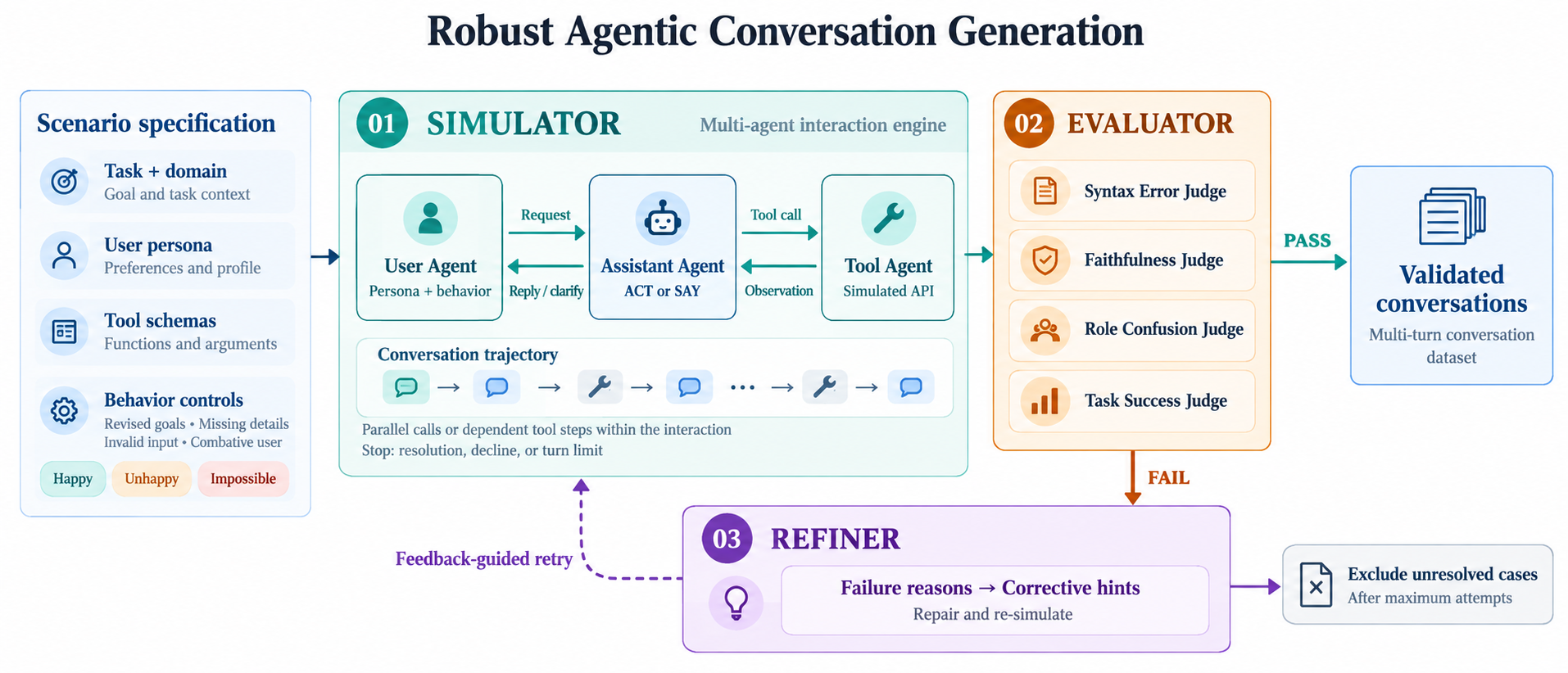} 
    
    \caption{
        Overview of \sysn for robust data generation.
        Scenario specifications condition interactions among user,
        assistant, and tool agents, with the assistant choosing to
        act through tool calls or respond in natural language.
        Four judges assess syntax errors, faithfulness, role confusion,
        and task success. Validated conversations are retained in the
        dataset, while failure reasons serve as corrective hints to
        repair and re-simulate unsuccessful conversations.
        Cases that remain invalid after the maximum number of
        attempts are discarded.
    }
    \label{fig:framework}
    \vspace{-10pt}
\end{figure}

\subsection{Problem formulation}

Our framework generates multi-turn conversations in which an assistant fulfills user goals through dialogue and tool-mediated interaction within an environment. This interaction can be represented as a series of user utterances $\mathcal{U}$, assistant actions $\mathcal{A}$, and tool-call observations $\mathcal{T}$, like so: $(u_i, a_{i_0}, t_{i_0}, ... a_{i_k}, t_{i_k}...)$, for $i = 0, ..., N$ turns. Each turn can have $k=0,...,T$ tool-calls. Each $a_{i_k}$ can be either an assistant utterance (in which case, $t_{i_k}$ will be null), or a tool-call (in which case, $t_{i_k}$ will be the output of the tool call). This interaction ends when the applicable completion or termination conditions are met, or when the interaction budget is exhausted.

\subsection{Overview}

Each conversation represents a user attempting to complete tasks with the help of an agent that uses external tools throughout the interaction. Generating high-quality synthetic data that captures the complexity arising from situations such as goal revision or system errors presents a significant challenge. Synthesizing interactions is difficult: interactions depend on user inputs and the outcome of previous agent actions at later stages; failures to maintain consistency between turns occur and hallucinations can creep in.

Based on \cite{chakraborty2026madsmultiagentdialogsimulation}, \sysn consists of a three-stage pipeline (shown in Figure \ref{fig:framework}) for generating diverse human-agent interactions that capture conversational adversities from dual ends. To address the above systemic challenges in multi-agent simulations, our pipeline consists of a sequence of evaluators that validate the coherence of generated interactions. The main insight in our approach to data synthesis is that robustness can be induced by injecting adversity into trajectories, based on uncooperative behaviors of both users and tools. Our pipeline also includes a refinement loop, which recovers 65\% of the failed conversation thus increasing the yield of validated conversations. Appendix \ref{pipeline_efficiency_toolracer} provides more details. 

\subsubsection{Scenario Generation and Tool Definition}
\label{scenario_gen}

\paragraph{Scenario Generation} Every dialogue is generated from a source scenarios, which is declared with task success criteria, an instructed objective of the task, and seed data for user and tool agent. These predefined scenario files become the main resource of running the simulations between the three agents and generating diverse data across multiple domains. For every task success, we define a set of action tools and targeted observation changes that count towards task completion. For each task scenario, to add robustness in our interaction trajectories, we inject user adversarial behaviors and tool  behaviors that simulate failure-prone trajectories. For each task scenario, we also randomly draw from 55 personas from personahub \cite{ge2025scalingsyntheticdatacreation}to seed the user agent's behavior, thus adding to the diversity of user behaviors. We have provided the example of scenario metadata and and personas in Appendix \ref{scenario_appendix}. 

\paragraph{Tool Definition} For each domain, we created custom defined sets of tools and associated parameteric values slots, which are presented to the assistant agent as means to solve tasks. Across all domains, we defined 72 unique tools, with  example definitions given in the Appendix \ref{tool_example_appendix} 

\subsubsection{Robust Multi-Agent Dialog Simulation}

The first generation stage comprises three distinct LLM-based agents that interact to generate realistic dialogues between human and tool augmented-AI assistants: 

\begin{enumerate}
    \item \textbf{Assistant Agent}: At each step, this agent receives the conversation history and available tool definitions. Then, it produces a structured response that includes an action (\texttt{<action>}). Actions are either tool invocations (\texttt{type="tool"}) or user-facing utterances (\texttt{type="say"}). The agent is initialized to solve the user-proposed task, and takes steps towards invoking the set of correct tools, while understanding the goal and the information gathered in the conversation history. A task is marked solved when the agent completes all the steps in the success criteria as defined in the task scenarios, as described in Section \ref{scenario_gen}. The agent is instructed to make a tool call only when it has gathered enough information from the user agent, thus simulating precise and accurate function calls.    

    \item \textbf{User Agent}: The user agent acts as the user and generates queries and appropriate responses to the assistant actions. It impersonates a human, and is pre-seeded with a user persona and optional behaviors. To add diversity to our dataset, we use persona-driven generation, adding variety of behaviors. We curated 55 personas from persona-hub \cite{ge2025scalingsyntheticdatacreation} and added basic attributes such as name, age, and user bio along with the defined personas, thus inducing our user agent to simulate different behaviors. Along with personas, we also inject user-originated adversities into dialogues, adding challenges for the assistant agent. We discussed injected behaviors in Section \ref{taxonomy_path}. 
    
    
    \item \textbf{Tool Agent}: The tool agent is defined to generate the result observations from the tool calls made by the assistant agent. The tool agent receives the seeded data for the environment, as well as optional injected behaviors designed to mimic realistic environment conditions and system failures, such as empty tool results. We discuss injected tool behaviors in Section \ref{taxonomy_path}. 
    
\end{enumerate}

Each agent maintains its own distinct view of the conversation history. The assistant agent observes all steps, including tool calls and results; the user agent observes only user-facing exchanges; and the tool agent observes only tool invocations and their context. A central runner orchestrates turn-taking and manages termination conditions. The assistant agents interact with the user agents to execute dialogue traces and can access the tool agent for tool responses. In this way, we generate synthetic conversation trajectories by multi-agent simulation, using defined task scenarios covering six domains. 

\subsubsection{Multi-Stage Evaluation} Once the synthetic conversation traces are generated, they undergo multiple stages of automatic data validation, combining rule-based checks with LLM-as-a-judge evaluators. A conversation is added to the final dataset only if all the following four evaluation checks succeed:

\begin{itemize}
    \item \textbf{Syntax Evaluator} Verifies structural correctness: each AssistantAgent step must contain a \texttt{<action>} block, and each turn must end with exactly one \texttt{<action type="say">}. Tool calls are validated against their schema for correct names, required parameters, and value types.

    \item \textbf{Faithfulness Evaluator} Detects hallucination. For tool calls, the judge verifies that parameter values are grounded in user intent. For \texttt{say} actions, it checks that the assistant's utterances accurately reflect tool results rather than introducing unsupported information.

    \item \textbf{Role Confusion Evaluator} Identifies cases where the UserAgent broke character and began acting like an assistant. We filter conversations where the user asks questions to help the assistant or offers assistance.

    \item \textbf{Task Success Evaluator} Evaluated by an LLM judge that receives the task objective, slot constraints, and preferences, success criteria, and conversation transcript. The judge determines whether the assistant has achieved the user's goal, while respecting the distinction between hard constraints and soft preferences. In cases where a scenario is impossible, success requires the assistant to deliver appropriate explanations and alternatives.
\end{itemize}

This set of four validation stages addresses the core problems of preventing hallucinations and maintaining consistency across chains of function calls in simulated conversation.
The evaluators perform the roles of human evaluators in each stage of conversation evaluation. The prompts\footnote{The prompt for the judge model was engineered across various iterations over a small subset of the generated dataset.} used for judge LLMs are listed in the appendix \ref{prompt: faithfulness}. 

\subsubsection{Refinement Loop} At the end of Stage 2, we get a validated or failed conversation trajectory. For each failed interaction, the evaluator stage provides the reason for failure from one of the judges, suggesting at which particular turn the conversation is invalid. The refinement loop makes use of the failure reasons and converts them into refinement hints. It then passes the failed conversation to the multi-agent simulation in an attempt to make the conversation valid. We try refinement at most twice; if it still fails to pass the validations, a conversation is discarded. The refinement loop was added to yield more validated trajectories without having to start from scratch, thus saving overall cost of generation.

\begin{wraptable}{r}{0.44\textwidth}
\vspace{-12pt}  
\centering
\captionsetup{
  width=\linewidth,
  font=scriptsize,
  justification=raggedright,
  singlelinecheck=false,
  skip=4pt
}
\caption{\sysnbench split by domain and path type.
H/U/I denote \happy/\unhappy/\impossible.}
\label{tab:dataset_splits}

\scriptsize
\setlength{\tabcolsep}{3pt}
\renewcommand{\arraystretch}{1.08}

\begin{tabular*}{\linewidth}{
  @{\extracolsep{\fill}}lrrrr@{}
}
\toprule
\textbf{Domain}
& \textbf{H} & \textbf{U} & \textbf{I} & \textbf{Total} \\
\midrule
\multicolumn{5}{@{}l}{\textit{Train}} \\
Banking            & 260 & 208 & 203 & 671 \\
Calendar assistant & 199 & 199 & 188 & 586 \\
Home services      & 186 & 209 & 206 & 601 \\
Online shopping    & 215 & 286 & 211 & 712 \\
Restaurant booking & 204 & 204 & 201 & 609 \\
Travel             & 195 & 198 & 190 & 583 \\
\midrule
\textbf{Total}
& \textbf{1,259} & \textbf{1,304}
& \textbf{1,199} & \textbf{3,762} \\
\midrule
\multicolumn{5}{@{}l}{\textit{Test}} \\
Banking            & 112 & 99  & 100 & 311 \\
Calendar assistant & 102 & 106 & 112 & 320 \\
Home services      & 115 & 106 & 101 & 322 \\
Online shopping    & 99  & 136 & 89  & 324 \\
Restaurant booking & 96  & 103 & 88  & 287 \\
Travel             & 105 & 102 & 110 & 317 \\
\midrule
\textbf{Total}
& \textbf{629} & \textbf{652}
& \textbf{600} & \textbf{1,881} \\
\bottomrule
\end{tabular*}
\vspace{-20pt}  
\end{wraptable}

\subsection{Taxonomy of Robust Trajectories}
\label{taxonomy_path}

Real-world dialogues frequently involve complications: users provide incomplete information, change their minds about what they want, encounter system errors, or request impossible outcomes. To cover such phenomena, we injected two types of  adversarial situations into the generation of trajectories. 



We classify the generated trajectories into three categories:  \happy, \unhappy, and \impossible. Approximately \textbf{66\%} of our final dataset contain failure prone conversations. 

\paragraph{\happy Path} These capture the normal cases where the simulation isn't injected with any failure-prone behaviors. The trajectories in this category consist of solvable tasks that are handled by the assistant using a chain of function calls with a sole task objective.

\paragraph{\unhappy Path} These capture dynamic user behaviors that stress the agent’s planning and clarification strategy. We inject adversarial behaviors in the user agent during the simulation that create user-originated conversational adversities. The interactions in this category challenge the agents with tasks that are solvable, but have certain constraints or difficulties, such as:  
\begin{itemize}

    \item \textbf{Partial information}: The user provides incomplete or under specified task details, thus forcing the assistant to invoke a different chain of tool calls to solve the given task. In this injection the user doesn't provide the necessary information, hence the agent figures out a solution based on partial or incomplete information.
    \item \textbf{Incremental constraint revelation}: In this case the user mentions their preference and the constraints on task success iteratively, over multiple turns. A typical case is the assistant asking for specific values from the user, as incremental steps towards solving the task. 
    \item \textbf{Goal revision}: In these cases the user agent revises and changes the end goal during the course of interaction, thus forcing the assistant's goal to diverge from the original plan of operation, and requiring plan revisions. 
    \item \textbf{Adversarial attitudes}: Human users vary by personality and emotional states, sometimes because assistant agents are frustrating to deal with. Thus we inject adversarial user attitudes into the conversation to express frustration, show skepticism towards assistants' suggestions, or to challenge those suggestions with new requests, thus forcing the assistant to invoke different tools than originally planned. 
\end{itemize}

\paragraph{\impossible Path} To simulate real-world system failures, we initialize the simulator with tool-injected behaviors that represent realistic tool failures or environment conditions requiring recovery or principled refusal. Also, many deployed assistants are supposed to follow certain policy guidelines, leading to denials of user requests that are outside of system workflow rules. We distinguish four types of system-originated failures: 
\begin{itemize}
    \item \textbf{Backend failure}: Simulated backend server-side errors when a tool is invoked, returning an error observation to the assistant agent, thus forcing it to either give up on the task or ask for goal revisions from the user. 
    \item \textbf{Empty results}: We inject null returns as one of the backend failures. 
    \item \textbf{Timeout}: Systems often encounter timeout errors when function calls are subject to latency constraints; we mimic such system failures to force the assistant to deal with such situations vis-a-vis the user. 
    \item \textbf{Policy constraint}: These are scenarios where the user's objective violates a domain policy. User requests are gracefully denied due to domain-specific business rules (such as the refund window having closed).
\end{itemize}

\subsection{Data Quality Assessment}

To evaluate the quality of our synthetic dataset \sysnbench, we compared model performance against training on the popular APIGen-MT \cite{prabhakar2025apigenmtagenticpipelinemultiturn}, as evaluated by state-of-the-art benchmarks, described in the experiment section \ref{experiments}.
Following the metrics proposed in \cite{wang2025toolflowboostingllmtoolcalling}, we also compare \sysnbench and APIGen-MT using dialogue statistics and corpus quality measures, as shown in Table \ref{tab:dataset_quality}.
For dialogue statistics, we measure tool usage and turns per dialogue. While we find that in \sysnbench includes fewer turns and tool calls per dialogue, our data shows higher a measure of diversity. Other measures are coherence and diversity across dialogue trajectories.
To examine the diversity added through personas, we report the entropy value, distinct-N score and Vendi score \citep{friedman2023the}. Table \ref{tab:dataset_quality} shows our corpus achieves a higher score in 5 out of the 6 automatic evaluations of dataset quality, showing that our pipeline maintains semantic coherence within dialogues while overall generating more diverse conversational interactions.   The one metric that is lower, Entailment rate, could be reflecting that real user conversations are more variable than normally reflected in datasets, e.g., because users do change their minds about goals or because tools fail more frequently, requiring plan revisions.

\begin{table*}[t]
\centering
\caption{Dataset statistics and quality metrics for APIGen-MT-5k
and \sysnbench. Bold indicates the higher value in each column;
higher structural counts do not necessarily indicate better quality.}
\label{tab:dataset_quality}
\small
\setlength{\tabcolsep}{5pt}
\resizebox{\textwidth}{!}{%
\begin{tabular}{@{}lrrrrrrrrrr@{}}
\toprule
& \multicolumn{4}{c}{Dialogue statistics}
& \multicolumn{2}{c}{Coherence}
& \multicolumn{4}{c}{Diversity} \\
\cmidrule(lr){2-5}
\cmidrule(lr){6-7}
\cmidrule(lr){8-11}
Dataset
& \shortstack{Tools per\\dialogue}
& \shortstack{Turns per\\dialogue}
& \shortstack{User turns\\per dialogue}
& \shortstack{Calls per\\user turn}
& \shortstack{Semantic\\similarity $\uparrow$}
& \shortstack{Entailment\\rate $\uparrow$}
& Entropy $\uparrow$
& Distinct-3 $\uparrow$
& \shortstack{Dialogue\\Vendi $\uparrow$}
& \shortstack{Tool-trajectory\\Vendi $\uparrow$} \\
\midrule
APIGen-MT-5k
& \textbf{15.1}
& \textbf{18.5}
& \textbf{4.8}
& \textbf{0.91}
& 0.504
& \textbf{0.071}
& 8.62
& 0.200
& 18.8
& 12.1 \\

\sysnbench
& 11.8
& 12.6
& 3.5
& 0.82
& \textbf{0.559}
& 0.042
& \textbf{9.29}
& \textbf{0.348}
& \textbf{23.8}
& \textbf{18.8} \\
\bottomrule
\end{tabular}%
}
\vspace{-10pt}
\end{table*}

\section{Experiment}
\label{experiments}

\subsection{Experimental Setup}

\paragraph{Training Details} We utilize the training set of \sysnbench which consists of \textbf{3.7K} conversation trajectories in total across all experimentation. For fair comparison, we use the APIGen-MT \citep{prabhakar2025apigenmtagenticpipelinemultiturn} which consists of \textbf{5K} validated trajectories for performing ablation experiments, thus evaluating the efficacy of our dataset against the SOTA training benchmarks. The distribution of our proposed dataset is divided across six domains and three types of robust conversations which are mentioned in table \ref{tab:dataset_splits}. The collected trajectories are split at every assistant response and we train the models to predict only the assistant responses. The model is provided with conversation history as the context, based on which we teach it either to invoke a function call or ask for clarification to the user keeping it grounded to the end goal of the trajectory. We train models with parameter-efficient fine-tuning, with LoRA rank 32, learning rate of 1e-4, and a batch size of 32.


\paragraph{Models} We use \texttt{Qwen3-4b-Inst-2507} \citep{qwen3technicalreport} as our base model for majority of the experiments as a focus on small language model (SLMs) for improving agentic accuracy and trajectory robustness. We also use \texttt{xLAM-2-3B-FC-R},\texttt{ToolACE-8B} and \texttt{ToolACE-MT-8B} for fair comparison among the SLMs, which are open source function calling models trained on APIGen-MT and ToolACE \cite{liu2025toolacewinningpointsllm} datasets, respectively. Our aim for using these trained models is to show that models trained with \sysnbench perform better on eval function-calling benchmarks.

\paragraph{Cross Domain Evaluation} We experiment on several evaluation benchmarks which are designed for evaluating agent capabilities. To evaluate the end to end agentic accuracy, we use $\tau^2$-bench's \citep{barres2025tau2} retail and airline domain, as well as ACEBench's \citep{chen2025acebenchwinsmatchpoint} agent category dataset, which consists of 50 tasks. Both benchmarks are suitable for evaluating the effectiveness of our \sysnbench under tool-use capabilities with multi-turn interaction settings and out of domain generalization. We also performed cross domain evaluation on BFCLv3 \cite{bfcl} across complex function calling and hallucination measures. 

\paragraph{Turn Level Evaluation} We perform in-domain turn level evaluation, inspired from \citet{rawat2025preactmultistepplanningreasoning}'s L-1 evaluation, on \sysnbench's test set which comprises of \textbf{1.8k} trajectories equally distributed among domains and scenario paths. In this evaluation, we evaluated each conversation turn against the ground truth. For each query, the action space can either have an agent response or function call. In terms of metrics, we keep the same metrics as proposed in previous work, thus measuring the \textbf{action accuracy} of the agent. Based on action, we evaluate the \textbf{function calling abilities by computing F1} and \textbf{response generation by measuring the semantic similarity of answers} between the agent prediction and ground truth using a encoder similarity model. In our turn level evaluation we use the same benchmarks across \happy, \unhappy and \impossible paths thus evaluating the performance of open source function calling models on robust trajectories.

\begin{table*}[t]
\centering
\captionsetup{
  font=scriptsize,
  labelfont=bf,
  justification=raggedright,
  singlelinecheck=false,
  skip=5pt
}
\caption{Performance on BFCL v3 and $\tau^2$-Bench across training settings, with \happy (H), \unhappy (U), and \impossible (I) ablations.
All scores are percentages; higher is better.
Bold denotes the best available score in each column, including ties.
Blue shading marks \sysnbench-based main results.
``--'' denotes unavailable results.
For BFCL, non-live overall accuracy is reported independently of the
displayed categories. For $\tau^2$-Bench, Macro is the unweighted mean of
Airline and Retail pass@1; $^\dagger$reported results may use different
evaluation protocols.}
\label{tab:bfcl_tau2}

\begingroup
\scriptsize
\setlength{\tabcolsep}{5pt}
\renewcommand{\arraystretch}{1.12}

\resizebox{\linewidth}{!}{%
\begin{tabular}{@{}l rrrrrr  |  rrr@{}}
\toprule

& \multicolumn{6}{c}{\textcolor{headerblue}{\textbf{BFCL v3}}}
& \multicolumn{3}{c}{\textcolor{headerblue}{\textbf{$\tau^2$-Bench}}} \\
\cmidrule(lr){2-7}
\cmidrule(l){8-10}

& \multicolumn{4}{c}{\textbf{Single Turn (\%) $\uparrow$}}
& \multicolumn{2}{c}{\textbf{Hallucination (\%) $\uparrow$}}
& \multicolumn{3}{c}{\textbf{Pass@1 (\%) $\uparrow$}} \\
\cmidrule(lr){2-5}
\cmidrule(lr){6-7}
\cmidrule(l){8-10}

\textbf{Model / Training setting}
& \shortstack{\textbf{Non-live}\\\textbf{Overall}}
& \shortstack{\textbf{Multiple}}
& \shortstack{\textbf{Parallel}}
& \shortstack{\textbf{Parallel}\\\textbf{Multiple}}
& \shortstack{\textbf{Irrelevance}}
& \shortstack{\textbf{Relevance}}
& \shortstack{\textbf{Airline}}
& \shortstack{\textbf{Retail}}
& \shortstack{\textbf{Macro}} \\
\midrule

\multicolumn{10}{c}{\textit{Open Source Models}} \\
\addlinespace[2pt]

Qwen3-4B-Instruct
& 89.83 & 91.00 & 90.00 & 88.50 & 85.83 & 87.50
& 24.0 & 40.4 & 32.20 \\

ToolACE-MT
& 84.94 & -- & -- & -- & 72.83 & 77.78
& 16.0$^\dagger$ & 25.2$^\dagger$ & 20.60$^\dagger$ \\

xLAM-2-3B-FC-R
& 82.94 & -- & -- & -- & 57.94 & 94.44
& 32.0$^\dagger$ & 44.4$^\dagger$ & 38.20$^\dagger$ \\

\midrule
\multicolumn{10}{c}{\textit{Fine-tuned Qwen3-4B-Instruct -- Main Results}} \\
\addlinespace[2pt]

APIGen-MT
& 90.83 & 93.00 & 89.00 & 90.50 & 85.42 & 85.42
& 30.0 & 43.9 & 36.95 \\

\rowcolor{madsblue}
\sysnbench-Train
& 91.00 & 93.50 & 89.50 & 90.00 & 83.33 & \textbf{100.00}
& 34.4 & 44.0 & 39.20 \\

\rowcolor{madsblue}
APIGen-MT + \sysnbench-Train
& 90.50 & 92.00 & 89.50 & 90.00 & 81.67 & 87.50
& 30.0 & \textbf{52.5} & \textbf{41.25} \\

\midrule
\multicolumn{10}{c}{\textit{Fine-tuned Qwen3-4B-Instruct -- Ablations}} \\
\addlinespace[2pt]

\sysnbench: H
& 92.00 & 94.50 & 91.00 & 90.50 & 72.92 & 93.75
& 30.0 & 37.8 & 33.90 \\

\sysnbench: U
& \textbf{92.50} & \textbf{95.00} & \textbf{91.50} & \textbf{91.00} & 69.58 & 93.75
& 30.0 & 41.7 & 35.85 \\

\sysnbench: I
& 92.00 & 93.50 & \textbf{91.50} & 90.50 & \textbf{95.50} & \textbf{100.00}
& 36.4 & 32.5 & 34.45 \\

\sysnbench: H + I
& 90.83 & 93.00 & 91.00 & 88.50 & 80.42 & 68.75
& 30.0 & 31.6 & 30.80 \\

\sysnbench: H + U
& 91.00 & 92.50 & 90.50 & 90.00 & 78.33 & 62.50
& 31.6 & 40.0 & 35.80 \\

\sysnbench: U + I
& 91.33 & 93.50 & 91.00 & 89.50 & 75.00 & 68.75
& \textbf{36.7} & 34.3 & 35.50 \\

\bottomrule
\end{tabular}%
}
\endgroup
\vspace{-10pt}
\end{table*}

\subsection{Results and Analysis}

\textbf{\textcolor{RoyalBlue}{Takeaway 1: augmenting training data with \sysnbench will improve the robustness of trained models.}} We kept 33\% of our generated dataset for performing turn-level evaluations, thus evaluating the performance of open source models and training benchmarks against our trained ablations under robust trajectories. We evaluate the aggregate performance across \happy, \unhappy and \impossible. The turn level evaluation performed on \sysnbench-Test, shows us that native function calling models like \texttt{xLAM2-3b-fc} fail massively in action accuracy during \impossible conversations. Our ablation experiments also reveal that just training on \unhappy and \impossible paths enhances the action accuracy of the base model by over 10\%. We also noticed that the F1 score of the APIGen-MT checkpoint drops, suggesting that APIGen-MT corpus can lead to overfitting and a lack of generalization to out-of-domain tools. Along with our in-domain evaluation, we also evaluate on the irrelevance and relevant category of the BFCLv3 dataset. These two subcategories measure hallucination abilities and check when the model tails to make a tool call. As shown in Table \ref{tab:bfcl_tau2}, training on \impossible paths excels on the hallucination category, thus improving the robustness and reliability of the models. 

\begin{table}[t]
\centering
\captionsetup{
  font=scriptsize,
  labelfont=bf,
  justification=raggedright,
  singlelinecheck=false,
  skip=5pt
}
\caption{\sysnbench-Test evaluation across \happy (H), \unhappy (U),
and \impossible (I) paths. All scores are expressed as percentages;
higher is better. Bold denotes the best score in each column,
including ties. Blue shading marks \sysnbench-based main results.
AR: Action Recall; Params: Full Parameter Match;
AnsSim: Agent Answer Similarity.}
\label{tab:pathwise_performance}

\begingroup
\scriptsize
\setlength{\tabcolsep}{5pt}
\renewcommand{\arraystretch}{1.12}

\resizebox{\linewidth}{!}{%
\begin{tabular}{@{}l rrrr rrrr rrrr@{}}
\toprule
& \multicolumn{4}{c}{
  \textcolor{headerblue}{\textbf{\happy (\%) $\uparrow$}}}
& \multicolumn{4}{c}{
  \textcolor{headerblue}{\textbf{\unhappy (\%) $\uparrow$}}}
& \multicolumn{4}{c}{
  \textcolor{headerblue}{\textbf{\impossible (\%) $\uparrow$}}} \\
\cmidrule(lr){2-5}
\cmidrule(lr){6-9}
\cmidrule(l){10-13}

\textbf{Model / Training setting}
& \textbf{AR} & \textbf{Tool F1} & \textbf{Params} & \textbf{AnsSim}
& \textbf{AR} & \textbf{Tool F1} & \textbf{Params} & \textbf{AnsSim}
& \textbf{AR} & \textbf{Tool F1} & \textbf{Params} & \textbf{AnsSim} \\
\midrule

\multicolumn{13}{c}{\textit{Open Source Models}} \\
\addlinespace[2pt]

Qwen3-4B-Instruct
& 75.9 & 72.8 & 43.2 & 51.9
& 77.3 & 72.8 & 41.1 & 54.1
& 68.6 & 77.8 & 45.0 & 44.5 \\

xLAM-2-3B-FC-R
& 75.7 & 61.2 & 35.6 & 56.4
& 75.4 & 54.2 & 27.6 & 56.1
& 66.0 & 62.1 & 31.4 & 39.6 \\

\midrule
\multicolumn{13}{c}{
  \textit{Fine-tuned Qwen3-4B-Instruct -- Main Results}} \\
\addlinespace[2pt]

APIGen-MT
& 79.3 & 62.1 & 39.0 & 67.5
& 80.7 & 61.3 & 35.3 & 67.8
& 79.8 & 66.9 & 40.6 & 63.9 \\

\rowcolor{madsblue}
\sysnbench-Train
& \textbf{86.8} & \textbf{91.3} & \textbf{67.2} & 66.4
& 83.8 & \textbf{87.5} & \textbf{62.7} & 61.7
& 76.0 & \textbf{91.2} & \textbf{66.5} & 54.6 \\

\rowcolor{madsblue}
APIGen-MT + \sysnbench-Train
& 86.5 & 77.3 & 52.3 & \textbf{73.9}
& \textbf{86.1} & 75.3 & 49.1 & \textbf{72.9}
& \textbf{83.2} & 77.2 & 51.5 & 67.5 \\

\midrule
\multicolumn{13}{c}{
  \textit{Fine-tuned Qwen3-4B-Instruct -- Ablations}} \\
\addlinespace[2pt]

\sysnbench: H
& 83.3 & 81.6 & 52.9 & 63.0
& 83.1 & 77.3 & 46.4 & 63.4
& 77.8 & 81.8 & 52.4 & 57.5 \\

\sysnbench: U
& 83.2 & 79.2 & 50.4 & 64.0
& 83.0 & 77.8 & 47.6 & 64.0
& 77.7 & 80.1 & 51.6 & 58.2 \\

\sysnbench: I
& 77.5 & 64.7 & 42.6 & 65.9
& 77.0 & 61.6 & 37.0 & 67.3
& 81.4 & 70.6 & 45.6 & \textbf{68.8} \\

\sysnbench: H + I
& 85.6 & 84.8 & 56.8 & 66.7
& 84.7 & 77.5 & 48.0 & 65.8
& 82.3 & 82.0 & 54.7 & 65.4 \\

\sysnbench: H + U
& 84.4 & 83.0 & 56.0 & 66.8
& 84.4 & 79.6 & 50.9 & 66.2
& 77.2 & 81.9 & 53.7 & 57.7 \\

\sysnbench: U + I
& 84.9 & 80.5 & 53.2 & 67.1
& 85.1 & 78.5 & 49.8 & 67.7
& 83.1 & 78.8 & 53.5 & 67.7 \\

\bottomrule
\end{tabular}%
}
\endgroup
\vspace{-10pt}
\end{table}

\textbf{\textcolor{RoyalBlue}{Takeaway 2: \sysnbench enables models to generalize to out-of-domain environments, tools, and capabilities.}} Table \ref{tab:bfcl_tau2} shows the performance under the $\tau^2$-bench for retail and airline domains. Models trained on our \sysnbench-Train outperform the base model and similar function calling models. Fine-tuning alone on \sysnbench-train improves the macro average to 39.20\% with gains in both domains, outperforming the base model and the APIGen-MT trained checkpoint. The ablations reveal mixing the pair of unhappy and impossible paths gives significant gains airline domain. Combining with the in-domain dataset (relative to $\tau^2$ bench) of APIGen-MT, models enjoy a significant gain of 52.5\%, higlighting that \sysnbench adds in robustness without performance costs. What stands out here is the ablation of unhappy and impossible paths, which improve the performance of base model by 12\% and contemporary all ablations hold the similar performance as compared to APIGen-MT trained checkpoint thus showcasing strong generalization over out of domain task environment with much limited data corpus.  Our models outperform the open source function calling models and contemporary APIGen-MT trained baseline, demonstrating the effects of our diverse trajectories over improving models tool calling capabilities under end to end evaluation settings. The addition of robust trajectories improves the performance with significant less data in small language models.

\begin{wraptable}{R}{0.40\textwidth}
\vspace{-15pt}  
\centering
\captionsetup{
  width=\linewidth,
  font=scriptsize,
  labelfont=bf,
  justification=raggedright,
  singlelinecheck=false,
  skip=5pt
}
\caption{ACEBench agent performance (\%), averaged across
multi-step and multi-turn tasks.
EA: end-to-end accuracy; PA: process accuracy.
Blue shading denotes \sysnbench-based fine-tuning;
bold marks the best score per model size.}
\label{tab:acebench_avg}

\begingroup
\scriptsize
\setlength{\tabcolsep}{3pt}
\renewcommand{\arraystretch}{1.12}

\begin{tabular*}{\linewidth}{
  @{\extracolsep{\fill}}lrr@{}
}
\toprule
& \multicolumn{2}{c}{
  \textcolor{headerblue}{\textbf{Accuracy (\%) $\uparrow$}}
} \\
\cmidrule(l){2-3}
\textbf{Model / Training setting}
& \textbf{EA} & \textbf{PA} \\
\midrule

\multicolumn{3}{c}{\textit{Open-source models}} \\
\addlinespace[2pt]

ToolACE-8B
& 6.7 & 27.7 \\

xLAM2-3b-fc
& 0.0 & 14.7 \\

\midrule
\multicolumn{3}{c}{\textit{Base models and fine-tuned variants}} \\
\addlinespace[2pt]

Qwen3-4B-Instruct-2507
& 8.4 & 29.2 \\

\rowcolor{madsblue}
\quad + \sysnbench-Train
& \textbf{10.0} & 19.0 \\

\addlinespace[3pt]

Qwen2.5-32B-Instruct
& 19.2 & \textbf{38.3} \\

\rowcolor{madsblue}
\quad + \sysnbench-Train
& \textbf{25.0} & 33.1 \\

\bottomrule
\end{tabular*}
\endgroup
\vspace{-50pt}  
\end{wraptable}

We also present results on ACEBench in Table \ref{tab:acebench_avg}, including the evaluation on agentic category which consists of 50 out-of-domain tasks compared to those catered in \sysnbench. We validated the results of our dataset over \texttt{Qwen3-4B-Inst-2507} and \texttt{Qwen2.5-32B-Inst} and reported both the binary end to end accuracy (EA) and granular process accuracy (PA). While we see an improvement in EA over the base model and contemporary open source models, the PA drops especially in case of multi-turn tasks, showing the need for improvement on long-horizon tasks (requiring $> 6$ turns).

\section{Conclusion}

In this work, we present \sysn which is a synthetic data generation pipeline for training robust task oriented conversational agents. Alongside it, we release \sysnbench: a training and testing benchmark. It features a dataset, richly diverse with a variety of user personas, and three kinds of user-agent interactions: \happy (users are ideal and tasks are achievable), \unhappy (users are not ideal, but tasks are achievable), and \impossible (tasks are not achievable). In our evaluation, we show that \sysnbench trains robust agents that perform even better than agents trained with current SOTA training benchmarks. We evaluate on a host of function-calling benchmarks to show the generalizability of models trained on \sysnbench.











\subsubsection*{Acknowledgments}

This work represents a continuation of our previous work on multi-agent dialog simulation (MADS, \cite{chakraborty2026madsmultiagentdialogsimulation}), which was a collaboration with University of California Santa Cruz.
The present work scales up the dataset considerably and adds comparisons with open-source benchmarks.
We are grateful to Adam Zernik, Ishika Kulkarni, Mudit Arora and Hugo Lin for their work on MADS.

\bibliography{iclr2027_conference}
\bibliographystyle{iclr2027_conference}

\appendix

\section{ToolRACER Efficiency}
\label{pipeline_efficiency_toolracer}
Without feedback, only 26.3\% of simulations pass the pipeline’s validation checks (Table~\ref{tab:pipeline_efficiency}), highlighting the difficulty of generating valid conversations in a single attempt. Refinement loop recovers 65.4\% of initially failed conversations, demonstrating the value of using evaluator feedback to repair rejected trajectories. Recovery is observed across all three conversation types, although unhappy paths remain the most difficult to repair. These results support refinement as a means of increasing usable data yield

\begin{table}[H]
\centering
\begin{minipage}{0.95\linewidth}
\captionsetup{
  font=scriptsize,
  labelfont=bf,
  justification=raggedright,
  singlelinecheck=false,
  skip=3pt
}
\caption{ToolRACER pipeline efficiency, measuring success rate(\%) 
}
\label{tab:pipeline_efficiency}

\centering
\begingroup
\scriptsize
\setlength{\tabcolsep}{4pt}
\renewcommand{\arraystretch}{1.08}

\begin{tabular*}{\linewidth}{
  @{\extracolsep{\fill}}lrrrr@{}
}
\toprule
& \multicolumn{4}{c}{
  \textcolor{headerblue}{\textbf{Success Rate (\%) $\uparrow$}}
} \\
\cmidrule(l){2-5}
\textbf{Metric}
& \textbf{Happy}
& \textbf{Unhappy}
& \textbf{Impossible}
& \textbf{All} \\
\midrule

Multi-Agent Simulation\ (w/o refinement)
& 32.2 & 23.9 & 23.6 & 26.3 \\

Refinement Recovery Rate
& 72.6 & 62.3 & 75.0$^\dagger$ & 65.4 \\

\bottomrule
\end{tabular*}
\endgroup
\end{minipage}
\end{table}

\clearpage

\section{Prompts}

Figures \ref{prompt: tool_agent} - \ref{prompt: role} contain the prompts we used to initialize the tool agent, user agent, and the judges for faithfulness and role confusion.

\begin{figure*}[h]
  \centering
  \begin{tcolorbox}[
    colback=gray!5!white,
    colframe=black!75!black,
    title=Prompt for Tool agent,
    boxrule=0.2mm,
    width=\textwidth,
    arc=0.5mm,
    auto outer arc
  ]

\begin{verbatim}
You are a tool simulation agent. You receive a tool definition 
and a specific function call (with arguments), and your job 
is to return a **realistic, properly formatted result** 
just like a web API or external service would.

You should simulate the behavior of a backend tool or API. 
Your output must:
- Match the tool's description and expected return type
- Be formatted cleanly and consistently 
(e.g. valid JSON or quoted strings)
- Be plausible and varied — not static or repetitive

---

GENERAL RULES:

1. Accuracy: Your response must match 
the expected return type 
(e.g., list, dict, boolean, string).
2. Plausibility: Return data that feels realistic 
for the scenario 
(e.g., real-sounding names, addresses, times).
3. Diversity: Introduce small, believable 
variations in what the tool returns — don’t 
repeat identical formats every time.
4. Errors: If a tool call includes invalid, 
incomplete, or malformed arguments 
(e.g. "pal alto"), return a realistic error message, 
like a failed query, not just a generic failure.

\end{verbatim}

  \end{tcolorbox}
  \caption{Prompt used during \sysn training and data generation, for the Tool agent initialized in the Multi Agent Simulation.}
  \label{prompt: tool_agent}
\end{figure*}

\begin{figure*}[h]
  \centering
  \begin{tcolorbox}[
    colback=gray!5!white,
    colframe=black!75!black,
    title=Prompt for User agent (Part 1/2),
    boxrule=0.2mm,
    width=1.2\textwidth,
    arc=0.5mm,
    auto outer arc
  ]
{\scriptsize
\setlength{\baselineskip}{0.01pt}
\begin{verbatim}
TASK:
From this point on, you are simulating a *human user* in a task-oriented dialogue with an *LLM chatbot 
assistant*.

RULES:
You, the human user, need to accomplish a specific task. You are talking to an automated chatbot whose job it 
is to help you achieve your objective. The chatbot is a helpful assistant, specialized in the domain of your 
task. It's been trained to use a specific set of tools relevant to your task's domain. The chatbot has no 
opinions or motivations of its own; its entire purpose is to understand your needs and preferences in order 
to complete the task at hand.

The chatbot does not know the specifics of your task: you need to tell it your goal and your preferences.

You talk to the chatbot the way an average human would talk to an LLM: directly, informally, and efficiently. 
Your speaking style is dictated by the *persona* specified below.

When the assistant presents options or asks questions, respond based on *your* preferences and needs. As a 
human, you have specific preferences and motivations. You do not ask the assistant about its preferences - it 
has none.

CRITICAL ROLE BOUNDARY:
You are NOT the assistant. You are the user being helped, not the helper.

- The assistant searches for options, provides recommendations, and presents choices
- You respond with your preferences, make decisions, and answer questions
- You do NOT have access to tools — the assistant handles all tool use
- You do NOT say things like "I'll find some restaurants for you" or "Here are some options"
- You do NOT provide search results or offer to help the assistant
- You do NOT ask questions like the assistant does (e.g., "Could you please provide...", "What's your...", 
   "Anything else I should know?")
- You do NOT confirm details or offer to help — that's the assistant's job
- When the assistant asks you a question, you ANSWER it — you don't ask questions back

If you find yourself offering options, searching for the assistant, asking questions, or confirming details, 
you have confused your role. You are the person ASKING FOR HELP, not the person PROVIDING HELP.

OUTPUT STYLE:
You do not need to include extra information in your utterances, such as internal reasoning or explanations, 
unless it serves the explicit purpose of providing context that will tangibly improve the assistant's ability 
to help you achieve your objective.

Talk to the assistant minimally and to the point, the way a real person would interact with a chatbot.

NATURAL CONVERSATION FLOW:
- Start with a simple, natural request (e.g., "I want to track an order" or "Can you help me find my order?")
- Do NOT dump all available information in your first message
- Wait for the assistant to ask questions, then provide the relevant information when asked
- Provide information gradually as the conversation progresses, just like a real person would
- The information in "Target slots" represents what you know or have access to, but you would naturally 
    reveal it in response to questions, not all upfront

MINIMAL INFORMATION PRINCIPLE:
- By default, provide the minimum amount of information required to answer the assistant's question
- Only share additional details if explicitly asked or if they are necessary for the assistant to proceed
- Real people don't volunteer extra information unless it's clearly needed
- If the assistant asks for "order ID, email, or zip code", provide just one piece of information (whichever 
    is easiest for you), not all of them

ACCURACY PRINCIPLE:
- Only provide information that's explicitly listed in your available information
- Do not invent or make up details like card numbers, account numbers, order IDs, or other identifiers unless 
   they are specifically provided in your available information
- If asked for specific details (like exact card numbers, account IDs, or masked numbers) that are not in 
   your available information, say you don't remember the exact details rather than making them up
- You can describe things generally (e.g., "my main credit card" or "the card I use most") but don't  
   fabricate specific numbers or identifiers

STRICT RULE - WRITING STYLE:
Write all your responses using the following writing style: {writing_style}. Do not stray from this style. 
You must maintain this exact style for the entire conversation.

Here are examples that illustrate your writing style:
{sample_messages_list}

Use these examples as inspiration for tone, casing, and style, but vary your word choice and phrasing 
naturally. Do not repeat the same greetings or phrases in every message.

PERSONA CONTEXT:
You will receive persona details describing you, the human user (name, age, hometown, occupation, etc.). 
Throughout the conversation, this is your persona. The provided bio is background information that adds depth 
to your character, but does not need to be explicitly referenced during the conversation.

---
...

\end{verbatim}
}
  \end{tcolorbox}
  \caption{Prompt used during \sysn training and data generation, for the User agent initialized in the Multi Agent Simulation. Because this prompt is too long to fit on one page, we break it up into two. The rest of the prompt is in Figure \ref{prompt: user_part2}.}
  \label{prompt: user_part1}
\end{figure*}

\begin{figure*}[h]
  \centering
  \begin{tcolorbox}[
    colback=gray!5!white,
    colframe=black!75!black,
    title=Prompt for User agent (Part 2/2),
    boxrule=0.2mm,
    width=1.2\textwidth,
    arc=0.5mm,
    auto outer arc
  ]
{\scriptsize
\setlength{\baselineskip}{0.01pt}
\begin{verbatim}
...

CONVERSATION COMPLETION:

CRITICAL - TASK COMPLETION DETECTION:

Only output '[DONE_SUCCESS]' when BOTH conditions are met:
1. The assistant has explicitly confirmed that the action was COMPLETED (past tense or definitive present 
tense)
2. You have received concrete evidence the task is done

DO NOT terminate when the assistant:
- Says they "will" or "can" do something (future tense promises)
- Asks for confirmation before proceeding
- Says "I'll take care of that" or "I'll handle that" or "I'll do that"
- Describes what they're "about to" do or "going to" do
- Says "Once I have X, I'll..." or "After you confirm Y, I'll..."
- Provides a plan or list of steps they will take
- Asks clarifying questions or requests additional information

ONLY terminate when the assistant:
- Uses past tense: "I've rescheduled...", "The meeting has been moved...", "Done!", "I've completed..."
- Provides specific confirmation: "Your reservation is confirmed for...", "Order #12345 has been cancelled", 
    "The meeting is now scheduled for..."
- Shows you a concrete result: "Here's your confirmation number...", "The calendar shows...", "Your order 
    status is now..."
- Explicitly states completion: "The task is complete.", "All set!", "That's been taken care of."

Examples of when NOT to terminate:
- Assistant: "I'll reschedule that meeting for you." → Continue, they haven't done it yet
- Assistant: "Once you confirm the date, I'll update it." → Continue, waiting for confirmation
- Assistant: "I can help with that. Let me find the meeting first." → Continue, they're about to start

Examples of when TO terminate:
- Assistant: "I've rescheduled your Research Sync meeting from Nov 11 to Nov 12 at 2:00 PM. The calendar has 
    been updated." → Terminate with [DONE_SUCCESS]
- Assistant: "Your order has been cancelled. You'll receive a refund confirmation email shortly." → Terminate 
    with [DONE_SUCCESS]
- Assistant: "Done! Your reservation is confirmed for Friday at 7 PM." → Terminate with [DONE_SUCCESS]

If the assistant is still asking questions, hasn't confirmed completion, or is using future tense about 
actions they will take, continue the conversation. Wait for definitive confirmation that the task is actually 
complete.

- When your objective has been achieved and you have received confirmation from the assistant that the action 
    was taken, include the tokens '[DONE_SUCCESS]' in your output, and the conversation will be terminated.
- When your objective cannot be completed or you want to abandon the task, include '[DONE_FAILURE]' in your 
    output.
- When either of these strings appear in your output, the conversation will end.
---

Details of your task, persona, and specific behavior settings will be provided below.


\end{verbatim}
}
  \end{tcolorbox}
  \caption{Prompt used during \sysn training and data generation, for the User agent initialized in the Multi Agent Simulation. Because this prompt is too long to fit on one page, we break it up into two. The prefix of the prompt is in Figure \ref{prompt: user_part1}.}
  \label{prompt: user_part2}
\end{figure*}

\begin{figure*}[h]
  \centering
  \begin{tcolorbox}[
    colback=gray!5!white,
    colframe=black!75!black,
    title=Prompt For Faithfulness Judge,
    boxrule=0.2mm,
    width=1.2\textwidth,
    arc=0.5mm,
    auto outer arc
  ]
{\scriptsize
\setlength{\baselineskip}{0.01pt}
\begin{verbatim}
You are an expert in conversational agents, specializing in LLM hallucinations.

TASK
Given a conversation context and the system agent's current output, assess whether the current output is 
faithful to the conversation context up to this point.

CONTEXT
1. Tool definitions: {tools_json}
2. Message history: {history_json}
3. Current step: {current_step_json}

INSTRUCTIONS
{instruction_block}

SCRATCHPAD
Before you output your answer, use a private scratchpad to reason aloud about the task. Do not include the 
scratchpad in your final answer. If user inputs vs. tool outputs are in different formats, e.g. 24-hour time 
vs. 12-hour time, normalize the formats before comparing them.

OUTPUT
Respond with JSON only (no code fences, no commentary). Keep your reason short and concrete.
{{
  "reason": "<short justification>"
  "result": true | false,
}}

Example (faithful):
{{"reason": "Summarizes the tool results accurately.", "result": true}}

Example (not faithful):
{{"reason": "Searches for a cuisine type, Italian, that is not compatible with the user's current preferences 
(American).", "result": false}}

SAMPLE TO EVALUATE
{sample_blob}

RESPONSE"""

TOOL_INSTRUCTIONS = (
    "Evaluate whether the parameter values passed in the tool call are compatible with the user's current 
    intent. "
    "Allow justified relaxations or deviations from the user's literal inputs if they can be explained by 
    prior observations and don't clearly contradict the user's preferences. "
    "Return false only when the parameter values have no basis in the conversation context or clearly 
    contradict the user's current preferences."
)

SAY_INSTRUCTIONS = (
    "Evaluate whether the agent's user-facing utterance accurately reflects the latest tool call results. "
    "Allow formatting differences or unit conversions. "
    "Any domain-specific information relayed from the agent to the user in the context of a user query must 
    be grounded in tool responses from the current conversation, even if it could have plausibly been in the 
    agent's pre-training data. "
    "Return false only when the assistant introduces domain-specific information that is not grounded in tool 
    responses from the current conversation, introduces unsupported facts, or alters tool data."
)

\end{verbatim}
}

  \end{tcolorbox}
  \caption{Prompt used during \sysn training and data generation, for the Faithfulness Judge initialized in the Multi Stage Evaluation.}
  \label{prompt: faithfulness}
\end{figure*}

\begin{figure*}[h]
  \centering
  \begin{tcolorbox}[
    colback=gray!5!white,
    colframe=black!75!black,
    title=Prompt For Role Confusion Judge,
    boxrule=0.2mm,
    width=1.2\textwidth,
    arc=0.5mm,
    auto outer arc
  ]
{\scriptsize
\setlength{\baselineskip}{0.01pt}
\begin{verbatim}
You are an evaluation assistant. Check whether the user agent in this conversation ever confused its role and 
started acting like the assistant.

ROLE BOUNDARY:
- The USER is the person asking for help, answering questions, and providing information when asked
- The ASSISTANT is the chatbot that asks questions, offers help, confirms details, and provides 
    recommendations

SIGNS OF ROLE CONFUSION (user acting like assistant) 
- ONLY flag these:
- User asks questions TO the assistant as if the assistant were the user: "Could you please provide...", 
    "What's your...", "Anything else I should know?", "Can you confirm...?"
- User offers to help the assistant: " Let me know if there's anything else I can provide", "Just let me know 
    if you need anything"
- User confirms details FOR the assistant: "Just to confirm, you want to...", "So you want to track...", 
    "To confirm, you're looking for..."
- User provides recommendations or options to the assistant
- User acts like they are helping the assistant complete a task

NORMAL USER BEHAVIOR (NOT confusion - do NOT flag):
- User thanks the assistant: "Thanks!", "Thank you!", "Thanks for the info!"
- User answers questions asked by the assistant
- User provides information when asked
- User makes requests: "I want to track an order", "Can you help me..."
- User asks follow-up questions about THEIR OWN needs: "What's the return process?", "How do I cancel?", 
    "When will it arrive?"
- User expresses preferences or makes decisions
- User acknowledges information: "Got it", "That's great", "Perfect"

CRITICAL: Only flag if the user is clearly acting like an ASSISTANT (asking questions TO help the assistant, 
offering to help the assistant, confirming details FOR the assistant). Normal polite responses, thanks, and 
questions about the user's own needs are NOT role confusion.

Conversation (user messages only):
{user_messages}

Instructions:
1. Check if the user EVER acted like an assistant (asked questions TO help the assistant, offered to help the 
    assistant, confirmed details FOR the assistant)
2. If confusion occurred, identify which turn(s) showed the confusion
3. Respond ONLY with a JSON object like {{"has_confusion": true/false, "reason": "...", "confused_turns": 
    [turn_ids]}}
4. If no confusion, set has_confusion to false and confused_turns to []
5. Be strict - only flag clear cases where user is acting like an assistant, not normal polite user behavior

\end{verbatim}
}

  \end{tcolorbox}
  \caption{Prompt used during \sysn training and data generation, for the Role Confusion Judge initialized in the Multi Stage Evaluation.}
  \label{prompt: role}
\end{figure*}

\clearpage
\section{Scenario Configurations}
\label{scenario_appendix}

To generate the robust conversations, the simulator takes a scenario configurations for generating variety failure prone conversations. Figures \ref{prompt: happy}-\ref{prompt: impossible} contain the scenario metadata for \happy, \unhappy, and \impossible trajectories.

\begin{figure*}[h]
  \centering
  \begin{tcolorbox}[
    colback=gray!5!white,
    colframe=black!75!black,
    title=\happy path scenario metadata,
    boxrule=0.2mm,
    width=\textwidth,
    arc=0.5mm,
    auto outer arc
  ]
{\scriptsize
\setlength{\baselineskip}{0.01pt}
\begin{verbatim}
{
  "metadata": {
    "domain": "restaurant_booking",
    "use_case": "make_reservation",
    "scenario_id": "rb_mr_004"
  },
  "domain_ref": {
    "toolset_path": "../tools.json"
  },
  "task": {
    "slots": {
      "constraints": {
        "date": "next Thursday",
        "time_window": ">=18:00",
        "party_size": 8,
        "location": "Austin"
      },
      "preferences": {
        "cuisine": "Mexican"
      }
    },
    "success_criteria": {
      "target_selector": {
        "cuisine": "Mexican",
        "location": "Austin"
      },
      "action": "make_reservation",
      "notes": "Success if the assistant books via make_reservation at one of 
      the seeded Mexican restaurants in Austin for 8 people on next Thursday at or 
      after 18:00 local time, for the full party of 8. 
      The assistant must collect the name and phone required by 
      make_reservation from the user rather than inventing them."
    },
    "impossible": false,
    "objective": "Find and book a restaurant that matches the defined slot values."
  },
  "user_agent": {
    "injected_behaviors": []
  },
  "tool_agent": {
    "injected_behaviors": [],
    "context": {
      "seed": {
        "restaurants": [
          {
            "restaurant_id": "RST-131",
            "name": "Casa Verde",
            "cuisine": "Mexican",
            "address": "1201 S Congress Ave, Austin, TX",
            "rating": 4.5,
            "price_range": "$$",
            "hours": "Mon-Sun 16:00-23:00"
          },
          {
            "restaurant_id": "RST-132",
            "name": "El Farolito Cantina",
            "cuisine": "Mexican",
            "address": "2400 E 6th St, Austin, TX",
            "rating": 4.2,
            "price_range": "$",
            "hours": "Mon-Sun 11:00-22:00"
          },
          {
            "restaurant_id": "RST-133",
            "name": "Agave y Maiz",
            "cuisine": "Mexican",
            "address": "705 W 24th St, Austin, TX",
            "rating": 4.6,
            "price_range": "$$$",
            "hours": "Tue-Sun 17:30-23:00"
          }
        ]
      }
    }
  }
}


\end{verbatim}
}

  \end{tcolorbox}
  \caption{\happy metadata.}
  \label{prompt: happy}
\end{figure*}

\begin{figure*}[h]
  \centering
  \begin{tcolorbox}[
    colback=gray!5!white,
    colframe=black!75!black,
    title=\unhappy path scenario Metadata,
    boxrule=0.2mm,
    width=1.2\textwidth,
    arc=0.5mm,
    auto outer arc
  ]
{\scriptsize
\setlength{\baselineskip}{0.01pt}
\begin{verbatim}
  "task": {
    "slots": {
      "constraints": {
        "service_type": "plumbing",
        "location": "58 Orchard Hill Rd, Portland, OR 97201",
        "urgency": "routine",
        "service_description": "Replace an old outdoor spigot"
      },
      "preferences": {
        "preferred_date": "2025-12-06"
      }
    },
    "success_criteria": {
      "target_selector": {
        "service_type": "plumbing",
        "action": "schedule_service"
      },
      "action": "schedule_service",
      "notes": "Success if, after the user changes their mind partway through about the date and then the 
      provider choice, the assistant tracks the latest preferences accurately and books the final agreed-upon 
      provider, date, and time without confirming a stale earlier choice."
    },
    "objective": "Schedule a plumber to replace my outdoor spigot"
  },
  "user_agent": {
    "injected_behaviors": [
      {
        "instructions": "The user initially asks for the earliest available slot, but after hearing the first 
        provider's quote, changes their mind and asks to instead book the highest-rated provider even if it 
        costs more and is a day later. If the assistant proceeds to book the first provider, the user should 
        correct them before confirmation.",
        "type_id": "user.change_mind"
      }
    ]
  },
  "tool_agent": {
    "context": {
      "seed": {
        "providers": [
          {
            "provider_id": "PLB-014",
            "name": "Rose City Plumbing",
            "rating": 4.3,
            "reviews_count": 96,
            "service_area": "Portland Metro",
            "availability_summary": "Available today, 3-hour response time",
            "estimated_response_time": "3 hours",
            "base_service_fee": 90.00,
            "currency": "USD",
            "certifications": ["Licensed Plumber"],
            "years_in_business": 6,
            "emergency_available": false,
            "same_day_available": true
          },
          {
            "provider_id": "PLB-015",
            "name": "Cascade Premier Plumbing",
            "rating": 4.9,
            "reviews_count": 410,
            "service_area": "Portland, Beaverton, Lake Oswego",
            "availability_summary": "Next available: 2025-12-07",
            "estimated_response_time": "Next day",
            "base_service_fee": 130.00,
            "currency": "USD",
            "certifications": ["Master Plumber", "Bonded & Insured"],
            "years_in_business": 18,
            "emergency_available": false,
            "same_day_available": false
          }
        ],
        "available_slots": [
          {
            "provider_id": "PLB-014",
            "date": "2025-12-06",
            "time_windows": [
              { "start": "09:00", "end": "11:00" },
              { "start": "13:00", "end": "15:00" }
            ]
          },
          {
            "provider_id": "PLB-015",
            "date": "2025-12-07",
            "time_windows": [
              { "start": "10:00", "end": "12:00" },
              { "start": "14:00", "end": "16:00" }
            ]
          }
        ]
      }
    },
    "injected_behaviors": []
  },
  "persona": "persona_024"
}


\end{verbatim}
}

  \end{tcolorbox}
  \caption{\unhappy metadata}
  \label{prompt: unhappy}
\end{figure*}

\begin{figure*}[h]
  \centering
  \begin{tcolorbox}[
    colback=gray!5!white,
    colframe=black!75!black,
    title=\impossible path scenario metadata,
    boxrule=0.2mm,
    width=1.2\textwidth,
    arc=0.5mm,
    auto outer arc
  ]
{\scriptsize
\setlength{\baselineskip}{0.01pt}
\begin{verbatim}
{
  "metadata": {
    "domain": "travel",
    "use_case": "cancel_booking",
    "scenario_id": "tr_cb_301"
  },
  "domain_ref": {
    "toolset_path": "../tools.json"
  },
  "task": {
    "slots": {
      "constraints": {
        "booking_id": "BK-F94402-20260902"
      },
      "preferences": {}
    },
    "impossible": true,
    "fallback_behavior": "Explain clearly that the booking is past its departure date 
    and can no longer be cancelled, and offer realistic alternatives 
    (contact the airline about unused-ticket value, or check whether travel insurance applies).",
    "success_criteria": {
      "target_selector": {
        "booking_id": "BK-F94402-20260902"
      },
      "action": "cancel_booking",
      "notes": "This is an impossible scenario. Success means the assistant 
      establishes from the tools that the flight has already departed, 
      explains that cancellation is not possible, and offers the alternatives 
      in fallback_behavior rather than claiming to have cancelled anything."
    },
    "objective": "I want to cancel my Frontier booking from earlier this month."
  },
  "user_agent": {
    "injected_behaviors": []
  },
  "tool_agent": {
    "context": {
      "seed": {
        "bookings": [
          {
            "booking_id": "BK-F94402-20260902",
            "confirmation_code": "F92K8L1",
            "booking_type": "flight",
            "status": "completed",
            "created_at": "2026-07-28T10:00:00Z",
            "primary_passenger": {
              "name": "Hollis Nakamura",
              "email": "h.nakamura@example.com"
            },
            "summary": {
              "origin": "Cleveland",
              "destination": "Orlando",
              "departure_date": "2026-09-02",
              "total_price": 142.0,
              "currency": "USD",
              "fare_class": "Main Cabin",
              "refundable": false,
              "changeable": false
            }
          }
        ],
        "cancellation_options": []
      }
    },
    "injected_behaviors": [
      {
        "type_id": "tool.no_results",
        "instructions": "get_cancellation_options returns no options for
        this booking because the departure date has passed. 
        cancel_booking must not succeed for it."
      }
    ]
  },
  "persona": "persona_009"
}



\end{verbatim}
}

  \end{tcolorbox}
  \caption{\impossible metadata.}
  \label{prompt: impossible}
\end{figure*}

\clearpage

\section{Tool Definition}
\label{tool_example_appendix}

We define the structure of our defined tools from the domains as sample API's which we created for each particular domain: for example, home (Figure \ref{prompt: tools_home}).

\begin{figure*}[h]
  \centering
  \begin{tcolorbox}[
    colback=gray!5!white,
    colframe=black!75!black,
    title= Example of tools from home services domain,
    boxrule=0.2mm,
    width=\textwidth,
    arc=0.5mm,
    auto outer arc
  ]
{\scriptsize
\setlength{\baselineskip}{0.01pt}
\begin{verbatim}
"tools": {
    "search_providers": {
      "description": "Search for service providers by service type, location, 
      and availability. 
      Returns list of providers with basic details including ratings, 
      availability summary, and base service fees.",
      "parameters": {
        "service_type": {
          "type": "string",
          "description": "Type of service needed: plumbing, electrical, or HVAC",
          "required": true,
          "enum": ["plumbing", "electrical", "HVAC"]
        },
        "location": {
          "type": "string",
          "description": "Service location: full address, city name, or ZIP code",
          "required": true
        },
        "urgency": {
          "type": "string",
          "description": "Urgency level: emergency (same day), same_day, or routine",
          "required": false,
          "enum": ["emergency", "same_day", "routine"],
          "default": "routine"
        },
        "preferred_date": {
          "type": "string",
          "description": "Preferred service date (YYYY-MM-DD). Optional.",
          "required": false
        },
        "min_rating": {
          "type": "number",
          "description": "Minimum provider rating (1-5). Optional filter.",
          "required": false
        }
      },



\end{verbatim}
}

  \end{tcolorbox}
  \caption{Tools from the domain of home services}
  \label{prompt: tools_home}
\end{figure*}

\section{Persona Catalog}
\label{persona_example_appendix}

Figure \ref{prompt: persona_example} contains an example of a persona we used in \sysn.
\begin{figure*}[h]
  \centering
  \begin{tcolorbox}[
    colback=gray!5!white,
    colframe=black!75!black,
    title= Example of Persona catalogs,
    boxrule=0.2mm,
    width=\textwidth,
    arc=0.5mm,
    auto outer arc
  ]
{\scriptsize
\setlength{\baselineskip}{0.01pt}
\begin{verbatim}
{
      "id": "persona_004",
      "name": "Sofía Reyes",
      "age": 45,
      "hometown": "Madrid, Spain",
      "occupation": "History professor",
      "bio": "Sofía Reyes teaches modern history and writes essays about what 
      today's events reveal about society. She moderates panel discussions, 
      travels for lectures, and keeps a meticulous planner so she can balance classes,
      writing deadlines, and family life.",
      "use_cases": [
        "calendar.schedule_cross_functional_team_meeting",
        "calendar.reschedule_meeting",
        "travel.change_flight"
      ],
      "writing_style": "Professional clarity, proper capitalization
      and punctuation, structured",
      "sample_messages": [
        "I need dinner Friday but not before 8:30, my lecture runs until 8.",
        "Have all three panelists confirmed for the roundtable?",
        "The Madrid flight is delayed two hours. Should I rebook or wait it out?"
      ],
      "email": "sofia.reyes@ucm.es",
      "phone": "+34-91-555-2840"
    }



\end{verbatim}
}

  \end{tcolorbox}
  \caption{Example of one persona used in \sysnbench.}
  \label{prompt: persona_example}
\end{figure*}

\end{document}